# MAR-MAER: Metric-Aware and Ambiguity-Adaptive Autoregressive Image Generation

Dong Kai[1, a], Bai Tingting[2,b]

[1]Lanzhou Vocational Technical College, China

[2] Lanzhou Vocational Technical College, China

[a]dongkai@lvu.edu.cn, [b]baitingting@lvu.edu.cn

Abstract. Autoregressive (AR) models have demonstrated significant success in the realm of text-to-image generation. However, they usually face two major challenges. Firstly, the generated images may not always meet the quality standards expected by humans. Furthermore, these models face difficulty when dealing with ambiguous prompts that could be interpreted in several valid ways. To address these issues, we introduce MAR-MAER, an innovative hierarchical autoregressive framework. It combines two main components. It is a metric-aware embedding regularization method. The other one is a probabilistic latent model used for handling ambiguous semantics. Our method utilizes a lightweight projection head, which is trained with an adaptive kernel regression loss function. This aligns the model's internal representations with human-preferred quality metrics, such as CLIPScore and HPSv2.As a result, the embedding space that is learned more accurately reflects human judgment. We are also introducing a conditional variational module. This approach incorporates an aspect of controlled randomness within the hierarchical token generation process. This capability allows the model to produce a diverse array of coherent images based on ambiguous or open-ended prompts. We conducted extensive experiments using COCO and a newly developed Ambiguous-Prompt Benchmark. The results show that MAR-MAER achieves excellent performance in both metric consistency and semantic flexibility. It exceeds the baseline Hi-MAR model's performance, showing an improvement of +1.6 in CLIPScore and +5.3 in HPSv2. For unclear inputs, it produces a notably wider range of outputs. These findings have been confirmed through both human evaluation and automated metrics.

1.Introduction

With the development of large-scale autoregressive (AR) model, the generation of text to image has made significant progress. However, there are still two main limitations. First, standard AR (autoregressive) training usually relies entirely on maximum likelihood estimation. It aims to optimize the accuracy of token level, but does not directly align the results with human centered evaluation indicators such as CLIPScore or Human Preference Score version 2 (HPSv2). Therefore, although the generated images may appear coherent in some parts, they still cannot accurately represent the expected hint. If we do not include feedback on specific indicators during the training, it will be very challenging to identify and correct this problem. Secondly, many existing models assume that the hints are clear and specific. In practice, prompts are usually vague, abstract or metaphorical.

In order to effectively solve the above problems, we propose MAR-MAER model. Firstly, the model adds the metric aware embedded regularization module (MAER), which calculates the metric value of low dimensional space and combines with the microscopic regression loss function to make the generated image more suitable for human preferences; Secondly, we add a conditional variational encoder to the random generation uncertainty of fuzzy prompts, and introduce a controllable random mechanism to provide a variety of reasonable explanations for fuzzy prompts. The experimental evaluation shows that the MAR-MAER model achieves good results in terms of measurement consistency and fuzzy interpretation flexibility.

## 2. Related Work

Diffusion models can get good results in image generation tasks. It makes images by slowly getting rid of noise in the data. Peeble[1] created a new diffusion model named DiT. It is built on the transformer architecture. This is a latent diffusion model for images. It uses a transformer instead of the common U-Net backbone. Bao[2] enhanced the transformer's cross - scale fusion feature ability and created the U-ViT model. Lu[3] proposed the FiT model. It's a transformer - based structure that can create images in any size and shape. Xiang[4] used DiT and added an image feature extractor. He also improved the way of fusing features. After that, he came up with a simple method for image control called VGen adapter (Vision Generalization adapter).This way made control accuracy much better.

The autoregressive method is one of the top ways to create images. Fluid[5] and GIVT[6] presented an autoregressive (AR) model. This model uses continuous markers to make the model better at expressing things. Li[7] created the MAR model. This model mixes diffusion models and autoregressive models. It forecasts each token based on diffusion loss. The images made look like those from diffusion models. Zheng[8] made the MAR model better and created the Hi-MAR model. This model uses low - res tokens as a guide. It adds a diffusion transformer head to finish image details. However, most of these methods still rely on VQ tokenizers. This may lead to loss of information and unstable generation. We build our work on this Hi-MAR level. We combine metric perceptual learning with fuzzy semantic cues to improve it.

To evaluate text-to-image models, people often use metrics that don't need reference images. These measures check how well the image and the text match. CLIPScore[9] and HPSv2[10] are popular choices. It figures out how similar the CLIP embeddings of the prompt and the generated image are. But it has a hard time getting details such as composition or how nice it looks. To solve this problem, we make the generator's embedding space match these human-centered scores during training.

## 3. Method.

Our approach includes three primary components: a hierarchical autoregressive backbone, a metric-aware regularization module (MAER), and an ambiguity-aware stochastic module. Figure 1 offers a comprehensive depiction of the entire operational workflow of our proposed MAR-MAER model. First, text prompts, such as the word "freedom," will be processed by the OpenCLIP text encoder. This encoder will transform these prompts into semantic embeddings. Afterward, these semantic embeddings are directed to two primary modules for further processing.

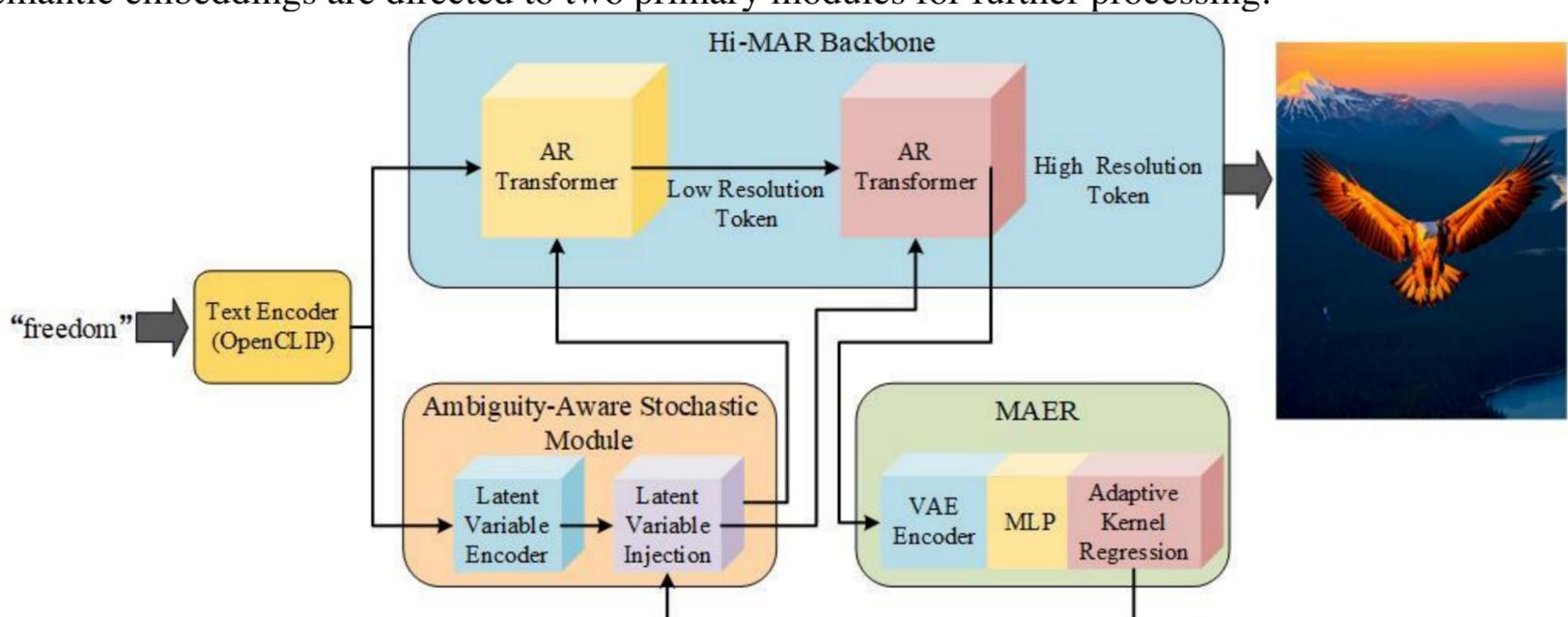

Figure 1: Overview of the MAR-MAER Architecture

The Ambiguity-Aware Stochastic Module randomly generates a latent variable C, which captures the various interpretations that the input text might hold. The latent variable C will be integrated into the Hi Mar backbone model through a two-step process. Initially, it is introduced as a prefix marker in the low-resolution autoregressive (AR) converter. Subsequently, in the second step, it is applied using the Feature-wise Linear Modulation technique during the high-resolution stage. This process guides the model to produce diverse and coherent image content.

The Measure Aware Embedded Regularization (MAER) module aims to improve the alignment between the model's output and human preference measurements. It uses an encoder, much like the one found in a variational autoencoder (VAE), to map the generated image to a latent space. This space utilizes adaptive kernel regression technology to ensure a smooth transition between the embedded similarity and the metric score, similar to CLIPScore. This feedback mechanism guarantees that the images generated by the model are not only visually realistic and persuasive, but also closely align with human aesthetic standards and semantic judgment.

3.1 Hierarchical Autoregressive Backbone

We utilize a two-stage autoregressive model following the Hi-MAR methodology. First, we encode the input text prompt T using a frozen CLIP text encoder. This enables us to generate a sequence of text embeddings $e_T$. Subsequently, we create the image through a two-step process:.

First stage, The model generates a low-resolution (LR) grid of visual tokens $x_{lr}$. These tokens capture the overall layout—like where objects are and what the scene looks like.

In the second stage, known as fine generation, the low-resolution (LR) tokens are upsampled and utilized as context to create a high-resolution (HR) token grid $x_{hr}$. This process adds fine details to the output. Both stages use standard autoregressive transformers. Here is the complete likelihood:

$$P(x_{hr}, x_{lr} | T) = P(x_{lr} | T) \cdot P(x_{hr} | x_{lr}, T) \qquad (1)$$

We train this with cross-entropy loss, denoted $L_{AR}$. This two-stage design helps the model focus on global structure first, then local details—just like humans do.

3.2 Metric-Aware Embedding Regularization (MAER)

The standard AR loss does not guarantee that the generated image will meaningfully correspond to the text for human observers. For example, the image might look realistic but could miss important details from the original prompt.

To resolve this issue, we introduce a new loss term called MAER. The model is developed to create images that match human-centered metrics like CLIPScore or HPSv2, utilizing their CLIP embeddings. For every generated image $I$, we calculate its CLIP image embedding $e_I$ utilizing a pre-trained and non-modifiable CLIP encoder. Furthermore, we compute a target score y—such as the CLIPScore between the image I and the text prompt T. To obtain a 2D point z, we utilize $e_I$, a small Multi-Layer Perceptron (MLP). In each training batch, we estimate the score $z_i$ by averaging the scores of its nearest neighbors, with these averages being weighted according to their respective distances (kernel regression). Our goal is to minimize the difference between the predicted score and the actual score. Formally, the MAER (Mean Absolute Error Rate) loss is defined as follows:

$$L_{MAER} = \frac{1}{N} \sum_{i=1}^{N} (\hat{y}_i - y_i)^2 \qquad (2)$$

where N is the number of observations, $y_i$ represents the actual value, and $\hat{y}_i$ is the predicted value. Importantly, the gradients from this loss are propagated back into the generator through the network. This adjusts the generator's internal representations so that nearby points in the embedding space receive similar quality scores, thus making the model's judgments more aligned with human assessments. MAER emphasizes local consistency instead of focusing solely on direct score regression. This approach makes it more stable and less prone to the influence of noisy labels.

3.3 Ambiguity-Aware Stochastic Module

Many prompts are unclear. For example, "a feeling of nostalgia" can evoke a range of emotions. A good model should be able to offer multiple valid interpretations instead of just one fixed response. To support this, we introduce a latent variable "c" that captures the intended meaning behind the prompt. We model c as a Gaussian random vector conditioned on the text: $c \sim N(\mu, diag(\sigma^2))$ where $\mu$ and $\sigma$ come from a small encoder that processes $e_T$. We incorporate c into both stages of generation. In the initial stage, we add "c" as a prefix to the text tokens. At test time, we can sample different c values to get diverse, yet coherent, images from the same prompt. During the second phase, we modify the upsampled LR features according to the context. This guarantees that the textures and colors align with the selected interpretation. To ensure that component C is properly accounted for during training, we incorporate a KL divergence term.

$$L_{KL} = KL(q(c|T) \| N(0,I))  \quad (3)$$

### 3.4 Training and Inference

We combine all losses into one objective:

$$L_{total} = L_{AR} + \lambda_1 L_{MAER} + \lambda_2 L_{KL}  \quad (4)$$

In our experiments, we defined $\lambda_1 = 0.05$ and $\lambda_2 = 10^{-4}$. During inference, users can select one output by utilizing the mean of the latent distribution. Generate multiple outputs by sampling different c values to achieve diversity. This makes our model flexible for both automatic and interactive use.

## 4 Experiments

We assess MAR-MAER using standard text-to-image benchmarks, along with several novel settings designed to evaluate metric alignment and ambiguity handling. Our experiments show that our model produces high-quality images, aligns more closely with human-preferred metrics compared to the baseline models, and demonstrates an impressive ability to generate diverse and meaningful outputs from vague prompts.

### 4.1 Experimental Setup

Available data sets include various categories and types that can be utilized to enhance research and analytics across multiple domains. They encompass fields such as healthcare, finance, environmental studies, social sciences, and more. Each dataset offers unique and valuable insights, supporting hypotheses, enabling predictive modeling, and facilitating data-driven decision-making. We use COCO Captions (118,000 images) for our training. The dataset includes over 10 million paired images and text. For evaluation, we use the COCO validation set, which includes 5,000 images, and we also introduce a new Ambiguous-Prompt Test Set. The implementation details are described in Table 1.

Table1:Implementation details

| Text encoder | frozen CLIP-ViT-L/14. |
|---|---|
| Optimizer | AdamW (lr = 1e−4, weight decay = 0.05) |
| Batch size | 1024 |
| Evaluation metric | FID、CLIPScore and HPSv2 |

### 4.2 Main Results

Table2: Quantitative results on COCO.

| Method | FID ↓ | CLIPScore ↑ | HPS v2 ↑ |
|---|---|---|---|
| Stable Diffusion | 12.1 | 27.8 | 16.2 |
| Fluid | 8.0 | 26.5 | 18.3 |
| MAR | 7.2 | 26.9 | 19.5 |
| Hi-MAR | 6.8 | 27.1 | 20.3 |
| MAR-MAER+ (Ours) | 6.3 | 28.7 | 25.6 |

We compare our results to those of robust autoregressive models. Table 2 shows that our model achieves the highest CLIPScore and HPS v2 compared to all other AR methods. Its FID score is also competitive—better than both Hi-MAR and Fluid, and nearly on par with diffusion models. We need to emphasize that we did not modify the AR architecture in any way. The improvements are solely due to MAER and the ambiguity module. This shows that both representation-level regularization and semantic modeling are essential, even when the architecture remains unchanged.

### 4.3 Ablation Study

Table3:Ablation experiment result.

| Model Variant | HPS v2 |
|---|---|
| Hi-MAR (baseline) | 20.3 |
| + MAER | 22.6 |

| + Ambiguity module | 21.2 |
| + MAER + Ambiguity (full model) | 25.6 |

The function of each part in MAR - MAER is shown in Table 3.Integrating only MAER can bring a big improvement. It's good to make the embedding space match human metrics, even if the generator's structure stays the same. The ambiguity module makes a small improvement. In general, MAR-MAER got good results.

4.4 Handling Ambiguous Prompts

To test ambiguity handling, we created a new test set of 200 abstract or multi-meaning prompts, such as: freedom, a digital memory, light in the darkness, the sound of silence.

For each prompt, we create images using various methods and ask 10 human raters to evaluate them based on two criteria.(1) Semantic Plausibility: Does the image present a believable and logical representation? (2) Diversity: Are the five outputs significantly different from one another in a meaningful manner? Our model excels in both these aspects.

Figure 2 provides examples of four different abstract prompts. Our model symbolizes "freedom" with an image of a bird flying high over the mountains. It looks open and unobstructed. In "A Digital Memory," a radiant city built entirely from data emerges into sight. It feels both futuristic and surreal. In the movie "Light in the Darkness," there is a single lantern casting a soft glow throughout an ancient library. Light shines brightly in the midst of darkness. When we think of "the sound of silence," we imagine a peaceful forest under the night sky. The moonlight shines through the branches of the trees. Silence filled the room. All four images fit the theme perfectly. This demonstrates that our model is capable of converting unclear words into accurate and varied images.

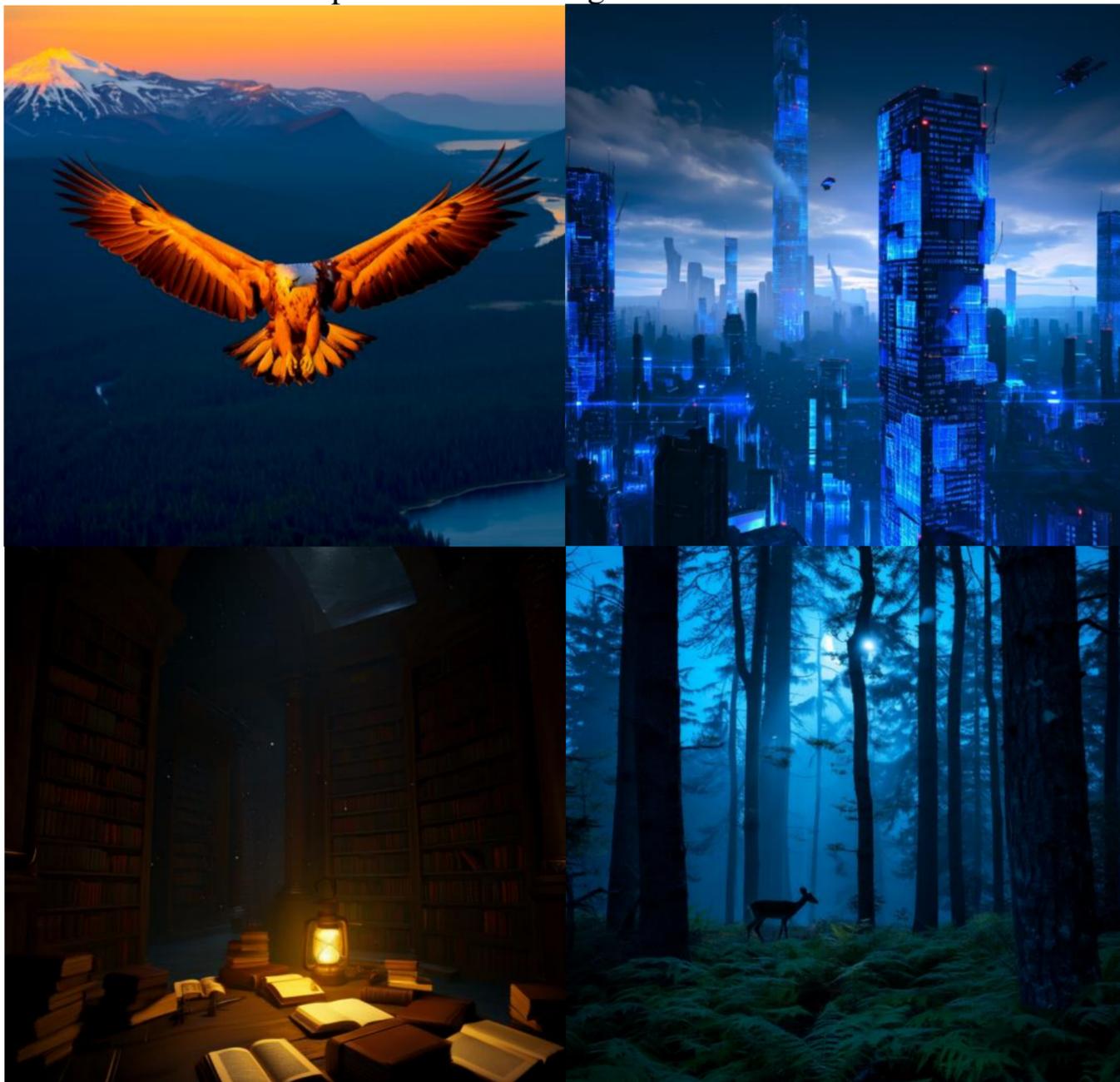

Figure2: Ambiguous prompts Generation

5 Discussion and Conclusion

Our research tackles two main issues present in existing autoregressive text-to-image models. Standard maximum likelihood training often fails to align well with human judgment of quality and meaning. For example, a model might interpret the term "light" solely as a physical phenomenon. To address this problem, we suggest using Metric-Aware Embedding Regularization. This method directs the generator to create images with CLIP embeddings that closely align with human-based metrics, such as CLIPScore and HPSv2. The result shows better alignment between the generated image and the prompt. Another issue is that deterministic models cannot address the inherent ambiguity present in abstract prompts. Our conditional variational module allows the model to explore different interpretations in a systematic way. For instance, "freedom" can be represented with images of a bird in flight, a vast, open landscape, or an empty, boundless space. Feedback from users indicates that our method yields results that are both more meaningful and more diverse.

In short, the MAR-MAER study shows that significant enhancements in alignment and flexibility can be achieved without changing the core architecture. Our approach leverages frozen CLIP features and relies on proxy metrics, offering a practical and adaptable framework. This framework aids in developing generative models that consider linguistic nuances and align them with human perceptual judgments.

6 Acknowledgements

This work was supported in part by the National College Student Innovation and Entrepreneurship Training Program of China under Grant No. 202412833015, and in part by the National College Student Innovation and Entrepreneurship Training Program of China under Grant No. 202412833001X.